\documentclass{article}
\usepackage{ijcai25}
\usepackage{times}
\usepackage{soul}
\usepackage{url}
\usepackage[hidelinks]{hyperref}
\usepackage[utf8]{inputenc}
\usepackage[small]{caption}
\usepackage{graphicx}
\usepackage{amsmath}
\usepackage{amsthm}
\usepackage{amsfonts}
\usepackage{amssymb}
\usepackage{booktabs}
\usepackage{algorithm}
\usepackage{algorithmic}
\usepackage[switch]{lineno}
\usepackage{color}
\usepackage{multicol}
\usepackage{multirow}
\usepackage[table]{xcolor}
\usepackage{colortbl}

\title{Exploring Transferable Homogeneous Groups for Compositional \\
Zero-Shot Learning}

\author{
    Zhijie Rao, Jingcai Guo\thanks{Jingcai Guo is the corresponding author.}, Miaoge Li, Yang Chen
    \affiliations
    Department of Computing, The Hong Kong Polytechnic University, Hong Kong SAR
    \emails
    \{zhijie96.rao, miaoge.li, cs-yang.chen\}@connect.polyu.hk, jc-jingcai.guo@polyu.edu.hk
}

\begin{document}

\maketitle

\begin{abstract}

Conditional dependency present one of the trickiest problems in Compositional Zero-Shot Learning, leading to significant property variations of the same state (object) across different objects (states). To address this problem, existing approaches often adopt either \textit{all-to-one} or \textit{one-to-one} representation paradigms. However, these extremes create an imbalance in the seesaw between transferability and discriminability, favoring one at the expense of the other. Comparatively, humans are adept at analogizing and reasoning in a hierarchical clustering manner, intuitively grouping categories with similar properties to form cohesive concepts. Motivated by this, we propose Homogeneous Group Representation Learning (HGRL), a new perspective formulates state (object) representation learning as multiple homogeneous sub-group representation learning. HGRL seeks to achieve a balance between semantic transferability and discriminability by adaptively discovering and aggregating categories with shared properties, learning distributed group centers that retain group-specific discriminative features. Our method integrates three core components designed to simultaneously enhance both the visual and prompt representation capabilities of the model. Extensive experiments on three benchmark datasets validate the effectiveness of our method.

\end{abstract}

\section{Introduction}

Deep neural networks have achieved impressive results in object recognition tasks, yet their capabilities remain underexplored in recognizing abstract concepts and performing compositional reasoning. Humans, on the other hand, possess a remarkable ability to decompose and reorganize underlying visual knowledge to recognize new concepts. For instance, after recognizing a \textit{red car} and a \textit{green truck}, one can easily identify a \textit{green car} and a \textit{red truck}, even without having seen these specific visual samples before. To exploit the compositional recognition potential of models, Compositional Zero-Shot Learning (CZSL)~\cite{chen2014inferring}\cite{misra2017red} has been proposed, aiming to generalize to unseen compositions by learning from seen state-object sample pairs.

Conditional dependency~\cite{wei2019adversarial}\cite{ge2022dual} is one of the most intractable problems in CZSL, which reflects the intrinsic connection between states and objects, often referred to as domain bias. The dependency results in the same state (or object) exhibiting vastly different visual features when paired with different objects (or states), making it difficult to generalize semantic information learned from seen compositions to unseen ones. To tackle this problem, existing approaches typically follow two fundamental paradigms, the \textit{all-to-one} and \textit{one-to-one} paradigms. The former focuses on learning a versatile, domain-invariant prototype, whereas the latter explicitly models the dependencies between objects and states to learn instance-specific representations. 

\begin{figure*}[t]
    \centering
    \begin{minipage}{\linewidth}
    \centering
    \includegraphics[width=0.99\textwidth]{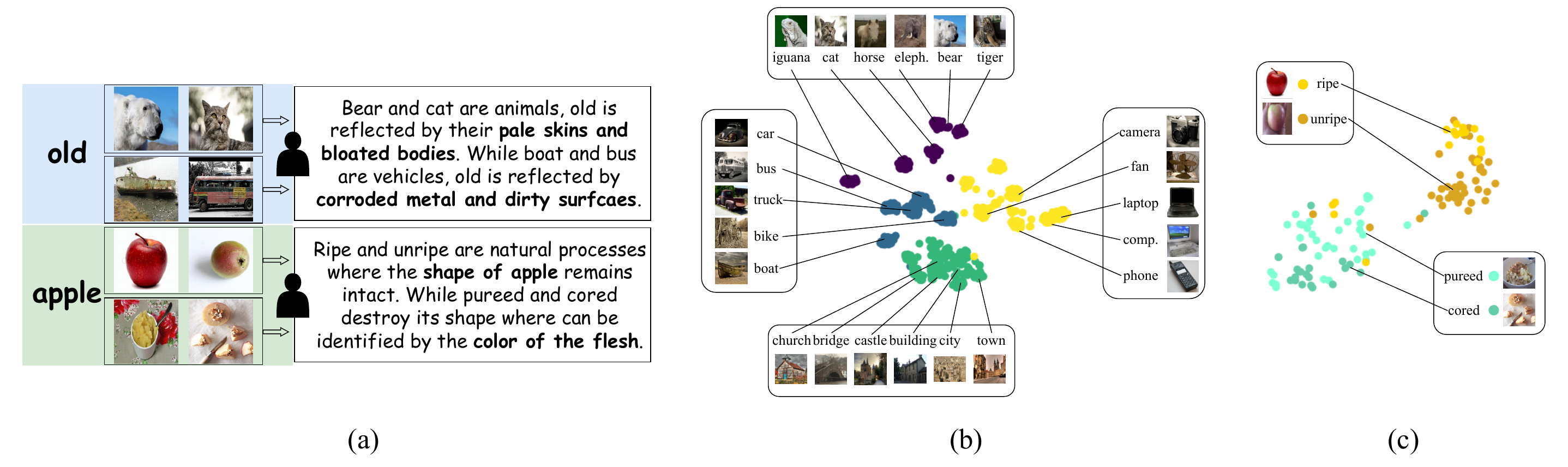}
    \end{minipage}
    \vspace{-2mm}
    \caption{(a) \textit{Motivation}. Humans utilize higher-order knowledge structure to perform hierarchical clustering, enabling effective analogies and reasoning. (b) \textit{Visualization of the state\textemdash old}. The semantic structure of homogeneous groups is well maintained in deep feature space, e.g., animals are naturally clustered together. (c) \textit{Visualizaiton of the object\textemdash apple}. Similarly, apples in various states cluster in groups due to visual variation.}
    \label{fig:intro}
\end{figure*}

Despite the promising results, it remains a challenge to strike a balance between transferability and discriminability~\cite{chen2019transferability}. The \textit{all-to-one} paradigm overly emphasizes transferability, aiming for a pure and uniform central representation. This focus forces the model to sacrifice instance-specific semantic information. In contrast, the \textit{one-to-one} paradigm prioritizes discriminability, leading the model to develop a bias toward the seen domain. Comparatively, humans are adept at analogizing and reasoning via hierarchical clustering to form group-specific conceptual cognition (Fig.~\ref{fig:intro} a). Such concepts maximize transferability among group members while preserving distinct and recognizable features. Thus, when recognizing an \textit{old dog}, we would easily associate it with a \textit{bear} or a \textit{cat} to reduce the complexity of inference. Fortunately, despite the lack of guidance by hierarchical labels in CZSL, we find that homogeneous semantic structures are naturally maintained in the deep feature space (Fig.~\ref{fig:intro} b-c).

Motivated by this, we propose Homogeneous Group Representation Learning (HGRL). Analogous to the idea of hierarchical clustering, we cluster categories with similar properties together and call it homogeneous group. The objective of HGRL is to identify unseen compositions by leveraging the high transferability among homogeneous group members. Our method integrates three critical components with CLIP as the backbone. Group-Aware Visual Representation (GAVR), aims to perceive and capture homogeneous groups and to learn group-specific representations to preserve semantic richness and completeness. To mimic the human knowledge system, we introduce the text co-occurrence probability graph to guide the hierarchical clustering process. Decoupled Group Prompt (DGP), a simple yet effective group-based prompt representation. Traditional prompts provide a centralized prototype for each category. However, finding a common prototype representation is extremely challenging. To this end, DGP equips each group with an exclusive context prompt to learn distributed prototype representations. Group-Aware Pair Enhancement (GAPE), which combines state and object features into pairs to facilitate joint recognition. To further improve the robustness and generalization of the model, we utilize the compatibility of homogeneous group members to enhance the state (object) features to obtain diverse pairs. We conduct experiments on three mainstream datasets and the results demonstrate that our method achieves state-of-the-art performance.

Our contributions are summarized below:

\begin{itemize}

    \item We analyze that the limitation of existing paradigms dealing with conditional dependency lies in the zero-sum game between transferability and discriminability. Inspired by the hierarchical clustering mechanism, we provide a new view to tackle this problem.

    \item We present a new framework, named HGRL, which integrates three essential components to enhance both visual and prompt representations. HGRL explores latent homogeneous groups and maintains group-specific semantic information with multi-expert networks and distributed prompts.

    \item We conduct experiments on three major benchmark datasets and the results show that the proposed method achieves state-of-the-art performance.
    
\end{itemize}

\section{Related Work}

\subsection{Compositional Zero-Shot Learning}

The goal of Compositional Zero-Shot Learning (CZSL) is to recognize unseen state-object pairs by learning and reorganizing seen compositions. The primary challenge is the distributional bias between seen and unseen compositions, which is caused by the dependency relationship between objects and states. Some studies adopt the \textit{all-to-one} paradigm, i.e., learning a versatile semantic representation for a single object or state. The dominant technique is decoupling representation, which is based on the idea that objects and states can exist and be represented independently, so that the two need to be divorced. To this end, prototype learning~\cite{ruis2021independent}\cite{lu2023decomposed} and attention-based decoupling techniques~\cite{saini2022disentangling}\cite{zhang2022learning}\cite{li2023distilled}\cite{hao2023learning} have been introduced. In addition, some studies are inspired by causal representations to uncouple the potential connection between objects and states through decorrelation techniques~\cite{atzmon2020causal}. Nan et al.~\cite{nan2019recognizing} introduce contrastive learning to compress the intra-class semantic space to purify the representations.

Another prominent paradigm is \textit{one-to-one} representation. The objective is explicitly modeling the dependency relations between states and objects. For example, CoT~\cite{kim2023hierarchical} uses object visual features as induced information to generate state visual features, while CANet~\cite{wang2023learning} and Troika~\cite{huang2024troika} leverage visual features to correct text embeddings. Other notable approaches include constructing conditional reconstruction networks~\cite{nagarajan2018attributes}\cite{li2020symmetry} and conditional classifiers~\cite{liu2023simple}\cite{huo2024procc}.


\noindent\textbf{Our contribution}: The \textit{all-to-one} and \textit{one-to-one} paradigms explore solutions to the issue of conditional dependency in almost opposing forms, yet both reach an extreme of neglecting discriminability or transferability. We provide a new perspective on it\textemdash find a balance between the two thus maximizing the generalization ability of the model.

\subsection{Hierarchical Clustering}

Hierarchical clustering is a classic clustering idea that considers both merging and splitting, drawing inspiration from tree-like human knowledge system~\cite{silla2011survey}. It plays a crucial role in various structured data tasks, such as text classification~\cite{chalkidis2020empirical}, speech classification~\cite{dekel2004online}, and protein function prediction~\cite{otero2009hierarchical}. Recently, hierarchical clustering has shown promising potential in vision tasks. For example, B-CNN~\cite{zhu2017b} combines the inter-layer outputs of convolutional networks with tree labels, significantly improving classification efficiency. Detclipv3~\cite{yao2024detclipv3} introduces hierarchical labels for open-vocabulary object detection, enabling the learning of multi-granularity semantic information. Additionally, some studies~\cite{chen2021hsva}\cite{novack2023chils} explore the potential of hierarchical clustering in zero-shot inference tasks.

\noindent\textbf{Our contribution}: To the best of our knowledge, we are the first to apply the idea of hierarchical clustering to CZSL to solve the problem of imbalance between discriminability and transferability. Meanwhile, we introduce the text co-occurrence probability graph to cope with the problem that no hierarchical labels are available in CZSL.

\section{Methodology}

\subsection{Preliminary}

\begin{figure*}[t]
    \centering
    \begin{minipage}{\linewidth}
    \centering
    \includegraphics[width=0.85\textwidth]{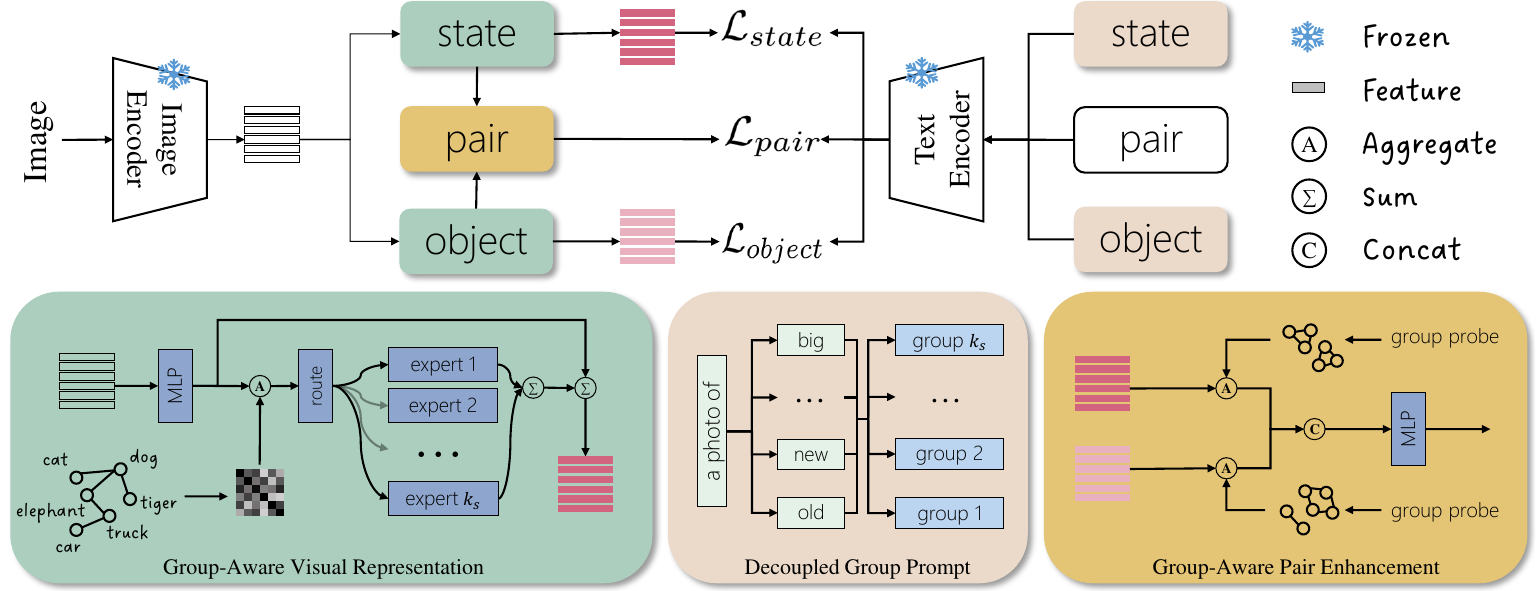}
    \end{minipage}
    \caption{Overview of the proposed method, which comprises three main components to enhance both visual and prompt representations. Note that the two branches of state and object are symmetric, so just one branch is presented. \textit{State (object) visual enhancement:} GAVR deeply explores latent homogeneous groups and utilizes multi-expert networks to extract group-specific representations to maintain semantic integrity and transferability. \textit{Pair visual enhancement:} GAPE integrates state and object features for joint recognition and introduces group-aware feature augmentation to improve pair diversity. \textit{Text prompt enhancement:} Unlike traditional category prompts, DGP additionally learns a customized contextual prompt for each group.}
    \label{fig:framework}
\end{figure*}

Suppose we have $n_s$ states $\mathcal{S}=\{s_1, s_2, ..., s_{n_s}\}$ and $n_o$ objects $\mathcal{O}=\{o_1, o_2, ..., o_{n_o}\}$, then there are $n_s\times n_o$ compositions defined as $\mathcal{C}=\{c_1=(s_1, o_1), c_2=(s_1, o_2), ..., C_{n_s\times n_o} = (s_{n_s}, o_{n_o})\}$, i.e., $\mathcal{C} = \mathcal{S} \times \mathcal{O}$. During the training phase, only partial compositions are available which are denoted as seen compositions $\mathcal{C}^{se}$. The goal of CZSL is to train a generalizable model thereby identifying unseen compositions $\mathcal{C}^{use}$. Note that $\mathcal{C}^{se}\bigcap\mathcal{C}^{use}=\emptyset$. Generally, $|\mathcal{C}^{se}|+|\mathcal{C}^{use}|<|\mathcal{C}|$ since some compositions in $\mathcal{C}$ do not exist in reality. So there are two settings in CZSL, closed-world CZSL where the label space is $\mathcal{C}^{se}\bigcup \mathcal{C}^{use}$ and open-world CZSL where the label space is $\mathcal{C}$. Formally, the training set is defined as $\mathcal{T}=\{(x_i, c_i)|x_i\in \mathcal{X}, c_i\in \mathcal{C}^{se}\}$, where $\mathcal{X}$ denotes the image space. An overview of our approach is presented in Fig. \ref{fig:framework}. 



\subsection{Group-Aware Visual Representation}

The representations learned by paradigm \textit{all-to-one} lose substantial key semantic information, while by paradigm \textit{one-to-one} tend to favor the seen domain. To solve this problem, we present Group-Aware Visual Representation (GAVR) to extract the state and object features, whose motivation is learning group-specific representations. The advantages is two-fold: 1) Maximizing semantic transferability among group memebers; 2) Preserving semantic integrity. \textbf{Since the two branches are symmetric, for ease of illustration, we next introduce the state branch only.}

The immediate problem is how to extract group-specific features. Inspired by mixture-of-expert network~\cite{jacobs1991adaptive}, we introduce multiple expert sub-networks to learn the features of different groups. Given an instance $(x, c) \in \mathcal{T}$, we obtain its semantic features via the image encoder $\Phi_I$. The feature $\Phi_I(x)$ at this point are a mixture of state and object semantic information. To get its state features, we use a sub-network $\Phi_S$ to map it to the state semantic space, which is denoted as $\Phi_S(\Phi_I(x))$. Previous methods would have used the feature to train directly, leading to a loss of group-specific information. We access a multi-expert network to extract group-specific semantic information. It consists of a route network $\Phi_{RS}$ and $k_s$ expert sub-networks $\Phi_E^S=\{\Phi_{E_1}^S, ..., \Phi_{E_{k_s}}^S\}$. First, the feature $\Phi_S(\Phi_I(x))$ is routed through the route network to determine the belonging of the group, the confidence for expert $i$ is denoted as:

\begin{equation}
    p_r^{S_i} = \frac{\exp(\Phi_{RS}(\Phi_S(\Phi_I(x)))_i)}{\sum_1^{k_s}\exp(\Phi_{RS}(\Phi_S(\Phi_I(x))))}.\label{Eq_route_confidence}
\end{equation}
We then assign features to the corresponding expert networks based on the route scores. To balance sparsity and smoothness, we use a top filtering mechanism, i.e., selecting $K$ experts with the largest confidence to participate in the next process. The obtained feature can be expressed as a weighted sum of $K$ experts:

\begin{equation}
    \mathbf{x}_g=\sum^K p_r^{S_i}\cdot \Phi_{E_i}^S(\Phi_S(\Phi_I(x))).
\end{equation}
Finally, we fuse the group-specific feature with the original feature:

\begin{equation}
    \mathbf{x}_s = \Phi_S(\Phi_I(x)) + \beta\cdot \mathbf{x}_g,
\end{equation}
where $\beta$ is a learnable parameter.

\textbf{How to Capture Homogenous Groups.} Another problem is the lack of supervision of hierarchical labels makes the capture of homogeneous groups uncontrollable, despite the natural heterogeneity exploration capability of route network. To this end, we utilize a word embedding model, GloVe~\cite{pennington2014glove}, to evaluate compatibility between categories. The core idea of GloVe is to learn co-occurrence relationships between words from a large-scale corpus. We argue that the probability of co-occurrence between words largely matches the probability that they belong to the same superclass in reality, and thus can be used as a substitute for hierarchical labels. For state branch, we need to evaluate the heterogeneity of different objects with the same state. For any two object words, we compute cosine similarity to assess the co-occurrence probability of them, which is denoted as $M^o_{ij}=\cos(\mathbf{v}^o_i, \mathbf{v}^o_j)$, where $\mathbf{v}$ denotes the word embedding extracted by GloVe.

Now the most straightforward approach is to assign hard group labels by clustering through similarity threshold filtering, which involves sensitive threshold effect and is prone to introduce noise. Inspired by graph representation learning~\cite{scarselli2008graph}, we encourage the model to adaptively discover homogeneous groups via local clustering. Specifically, we aggregate the feature of one sample with its nearest compatible samples. For example, for an \textit{old dog}, we fuse it with \textit{old cat}, \textit{old tiger}, \textit{etc.} to form a tiny homogeneous group. The model iteratively learns and summarizes the relationships of the tiny homogeneous groups to extend to the whole homogeneous group. Given a batch of instances $\{x_i, c_i=(s_i, o_i)\}^B \in \mathcal{T}$, where $B$ denotes the batch size. We can get their relation map $A\in \mathbb{R}^{B\times B}$, and we have:
\begin{equation}
    A_{ij}=M^o_{ij}\cdot\mathbb{I}[M^o_{ij}\geq\zeta]\cdot\mathbb{I}[s_i=s_j],\label{Eq_adjacency_matrix}
\end{equation}
where $\mathbb{I}[\cdot]$ denotes indicator function and $\zeta$ is a constant which is fixed to $0.5$. After normalization for $A$, we can get the aggregated features which denoted as
\begin{equation}
    \mathbf{x}^s_{agg}=A \otimes \Phi_S(\Phi_I(x)), A=\frac{\exp(A+\Lambda)}{\sum^{B}_{j=1}\exp(A_{ij}+\Lambda)}
\end{equation}
where $\otimes$ denotes matrix multiplication and $\Lambda$ denotes the unit diagonal matrix. Next the aggregated features are fed into route and expert networks for training.

\subsection{Decoupled Group Prompt}

Textual prompts are important components in visual language models (VLMs)~\cite{radford2021learning}. In the classification tasks, they act as primitive representations of categories thus providing optimized targets for visual features. The common practice of designing prompts in the fixed form of \textit{a photo of [class]}, which has been proven to perform inferior to learnable prompts~\cite{zhou2022learning}. The central idea of learnable prompts is to learn the task-specific context. To this end, previous CZSL approaches equip each branch with a separate contextual prompt, which are represented as:

\begin{equation}
    \mathbf{P}_i^S = [\mathbf{p}_1^s, ..., \mathbf{p}_m^s, \mathbf{w}^s_i],
\end{equation}
\begin{equation}
    \mathbf{P}_i^O = [\mathbf{p}_1^o, ..., \mathbf{p}_m^o, \mathbf{w}^o_i],
\end{equation}
\begin{equation}
    \mathbf{P}_i^C = [\mathbf{p}_1^c, ..., \mathbf{p}_m^c, \mathbf{w}^s_i, \mathbf{w}^o_i],\label{Eq_pair_prompt}
\end{equation}
where $\mathbf{P}_i^S, \mathbf{P}_i^O, \mathbf{P}_i^C$ denote the state, object and pair prompts, respectively. And $\{\mathbf{p}_1^s,...,\mathbf{p}_m^s\}$, $\{\mathbf{p}_1^o,...,\mathbf{p}_m^o\}$, $\{\mathbf{p}_1^c,...,\mathbf{p}_m^c\}$ denote their contextual prompts and $\mathbf{w}^s_i$, $\mathbf{w}^o_i$ denote the state and object words. For state and object branches, the learned prompts capture the global context yet ignore the discrepancies between heterogeneous groups. To this end, we add the corresponding prompt for each group, which are represented as:
\begin{equation}
    \mathbf{P}_i^{S_j} = [\mathbf{p}_1^s, ..., \mathbf{p}_m^s, \mathbf{g}_j^s, \mathbf{w}^s_i], j=1, 2, ..., k_s,
\end{equation}
\begin{equation}
    \mathbf{P}_i^{O_j} = [\mathbf{p}_1^o, ..., \mathbf{p}_m^o, \mathbf{g}_j^o, \mathbf{w}^o_i], j=1, 2, ..., k_o,
\end{equation}
where $k_s$, $k_o$ represent the groups of state and object, $\mathbf{g}^s$, $\mathbf{g}^o$ represent the group prompts. Then these prompts are fed into the text encoder $\Phi_T$ to obtain representations, which are denoted as:
\begin{equation}
    \mathbf{t}_i^{S_j}=\Phi_T(\mathbf{P}_i^{S_j}), \mathbf{t}_i^{O_j}=\Phi_T(\mathbf{P}_i^{O_j}).
\end{equation}

\textbf{Training Objectives.} We adopt InfoNCE loss function to train the network. At the same time, we use the confidence scores in Eq. (\ref{Eq_route_confidence}) as soft group labels for samples. For the state branch, the loss formula is:

\begin{equation}
    \mathcal{L}_{state}=-\frac{1}{B}\sum^B\log\frac{\sum_{j=1}^{k_s}p_r^{S_j}\cdot\exp(\mathbf{x}_s\cdot\mathbf{t}_i^{S_j}/\tau)}{\sum_{i=1}^{n_s}\sum_{j=1}^{k_s}\exp(\mathbf{x}_s\cdot\mathbf{t}_i^{S_j}/\tau)},
\end{equation}
where $\tau$ denotes the temperature coefficient. Since the object and state branches are symmetric, similarly, we can get the loss formula for the object branch as:
\begin{equation}
    \mathcal{L}_{object}=-\frac{1}{B}\sum^B\log\frac{\sum_{j=1}^{k_o}p_r^{O_j}\cdot\exp(\mathbf{x}_o\cdot\mathbf{t}_i^{O_j}/\tau)}{\sum_{i=1}^{n_o}\sum_{j=1}^{k_o}\exp(\mathbf{x}_o\cdot\mathbf{t}_i^{O_j}/\tau)}.
\end{equation}

\subsection{Group-Aware Pair Enhancement}

To improve the compositional recognition ability of the model, we introduce a pair branch. It utilizes the outputs of state and object branches as inputs to learn semantic reorganization and interaction. The pair branch consists of a learnable network $\Phi_P$. We concat the state features $\mathbf{x}_s$ and object features $\mathbf{x}_o$ as inputs, and the outputs are features with the same dimension as $\Phi_I(x)$. To enhance the performance of $\Phi_P$, we introduce group-aware feature enhancement to augment the input space. Although previous approaches have used similar enhancement techniques~\cite{panda2024compositional}\cite{jing2024retrieval}, the difference is that we argue that not all enhancements are meaningful due to the heterogeneity of groups. For example, it is inappropriate to enhance an \textit{old dog} with \textit{old} semantics extracted from an \textit{old car}. Therefore, we suggest that compatibility between instances should first be checked.

To realize this, we employ the group confidence scores in Eq. (\ref{Eq_route_confidence}) to detect homogeneous groups. For one instance, the group confidence scores of its state features can be denoted as $p_r^S(i)=[p_r^{S_1}(i), p_r^{S_2}(i), ..., p_r^{S_{k_s}}(i)]$. The compatibility of two instances with respect to the state can then be quantized as $\cos(p_r^S(i), p_r^S(j))$. Similar to Eq. (\ref{Eq_adjacency_matrix}), we can obtain an adjacency matrix $\hat{A^s}$:
\begin{equation}
    \hat{A^s}_{ij}=\cos(p_r^S(i), p_r^S(j))\cdot\mathbb{I}[\cos(p_r^S(i), p_r^S(j))\geq\zeta]\cdot\mathbb{I}[s_i=s_j].
\end{equation}
Then the enhanced state features are formulated as:
\begin{equation}
    \hat{\mathbf{x}}_s=\hat{A^s}\otimes \mathbf{x}_s, \hat{A^s}=\frac{\exp(\hat{A^s}+\Lambda)}{\sum^{B}_{j=1}\exp(\hat{A^s}_{ij}+\Lambda)}.
\end{equation}
Similarly, we can get the enhanced object features, denoted as $\hat{\mathbf{x}}_o$. Then we concat them to form the pair features, which are denoted as $\bar{\mathbf{x}}_p=\hat{\mathbf{x}}_s\oplus\hat{\mathbf{x}}_o$, where $\oplus$ means concat. Then we can get the output of pair branch and pair prompt representations according to Eq. (\ref{Eq_pair_prompt}):
\begin{equation}
    \mathbf{x}_p=\Phi_P(\bar{\mathbf{x}}_p), \mathbf{t}_i^{C}=\Phi_T(\mathbf{P}_i^{C}).
\end{equation}
The training loss is formulated as:
\begin{equation}
    \mathcal{L}_{pair}=-\frac{1}{B}\sum^B\log\frac{\exp(\mathbf{x}_p\cdot\mathbf{t}_i^{C}/\tau)}{\sum_{i=1}^{|\mathcal{C}^{se}|}\exp(\mathbf{x}_p\cdot\mathbf{t}_i^{C}/\tau)}.
\end{equation}

\subsection{Summarize}
\textbf{Total Training Loss.} The basic loss that applies CLIP to the CZSL task is denoted as:
\begin{equation}
    \mathcal{L}_{base}=-\frac{1}{B}\sum^B\log\frac{\exp(\Phi_I(x)\cdot\mathbf{t}_i^{C}/\tau)}{\sum_{i=1}^{|\mathcal{C}^{se}|}\exp(\Phi_I(x)\cdot\mathbf{t}_i^{C}/\tau)}.
\end{equation}
Then the total training loss of the propose method is denoted as:
\begin{equation}
    \mathcal{L}_{HGRL}=\mathcal{L}_{base}+\lambda\cdot(\mathcal{L}_{state}+\mathcal{L}_{object}+\mathcal{L}_{pair}), \label{Eq_total_loss}
\end{equation}
where $\lambda$ is a hyper-parameter.

\textbf{Inference.} We integrate the outputs of backbone, state, object, and pair branches for joint prediction. The output of backbone is represented as:
\begin{equation}
    p_{base}(c|\Phi_I(x))=\frac{\exp(\Phi_I(x)\cdot\mathbf{t}_i^{C}/\tau)}{\sum_{i=1}^{|\mathcal{C}^{se}|}\exp(\Phi_I(x)\cdot\mathbf{t}_i^{C}/\tau)}.
\end{equation}
Similarly, the output of pair branch is denoted as $p_{pair}(c|\mathbf{x}_p)$. For output of state branch, it is denoted as:
\begin{equation}
    p_{state}(s|\mathbf{x}_s)=\frac{\sum_{j=1}^{k_s}\cdot\exp(\mathbf{x}_s\cdot\mathbf{t}_i^{S_j}/\tau)}{\sum_{i=1}^{n_s}\sum_{j=1}^{k_s}\exp(\mathbf{x}_s\cdot\mathbf{t}_i^{S_j}/\tau)}.
\end{equation}
Similarly, we can get the output of object branch, denoted as $p_{object}(o|\mathbf{x}_o)$. The final prediction is:
\begin{equation}
    \mathop{\arg\max}_{c\in \mathcal{C}^{tar}} p_{base}+p_{pair}+p_{state}\times p_{object},
\end{equation}
where $\mathcal{C}^{tar}$ is $\mathcal{C}^{se}\bigcup\mathcal{C}^{use}$ for closed-world setting or $\mathcal{C}$ for open-world setting.

\subsection{Theoretical Insights}

We provide theoretical insights for our approach from the perspective of domain adaptation~\cite{ben2006analysis} since one state (object) on different objects (states) can be viewed as different domains~\cite{zhang2022learning}.

\noindent\textbf{Definition 1} \textit{Given a seen domain, $\mathcal{S}$, and an unseen domain, $\mathcal{U}$. We have,
\begin{equation}
    \epsilon_{U}(h)\leq\epsilon_{S}(h)+d(\mathcal{S}, \mathcal{U})+\gamma,
\end{equation}
where $\epsilon_{U}(h), \epsilon_{S}(h)$ is the expected risks for unseen and seen domains and $h$ is the model. $d(,)$ is the distributional divergence and $\gamma$ is a constant denotes the minimum of the risk sum of $h$ over $\mathcal{S}$ and $\mathcal{U}$.
}

Def. 1 reveals a loose upper bound which is determined by the distributional divergence between $\mathcal{S}$ and $\mathcal{U}$. Let $\mathcal{S}^\ast\in \mathcal{S}$ denotes the homogeneous group for $\mathcal{U}$ and we have $\mathcal{S}^\sharp=\mathcal{S}-\mathcal{S}^\ast$. Then $d(\mathcal{S}, \mathcal{U})$ could be decomposed into:
\begin{equation}
    d(\mathcal{S}^\ast, \mathcal{U}|p(\mathbf{g}_{u}=\mathbf{g}_{s^\ast}))+d(\mathcal{S}^\sharp, \mathcal{U}|p(\mathbf{g}_{u}\neq\mathbf{g}_{s^\ast})),
\end{equation}
where $\mathbf{g}$ denotes group label. Obviously $d(\mathcal{S}^\ast, \mathcal{U})\leq d(\mathcal{S}, \mathcal{U})$ and $d(\mathcal{S}^\ast, \mathcal{U})\leq d(\mathcal{S}^\sharp, \mathcal{U})$. To this end, our method first creates conditions for separating homogeneous group $\mathcal{S}^\ast$ by constructing a multi-expert network as well as distributed group prompts. Then, the text co-occurrence probability graph is utilized to guide the clustering of homogeneous categories thereby increasing the probability $p(\mathbf{g}_{u}=\mathbf{g}_{s^\ast})$.

\section{Experiments}

\subsection{Experiment Settings}

\begin{table*}[t]
    \centering
    \begin{tabular}{l | c | c c c c | c c c c | c c c c}
        \toprule
         & & \multicolumn{4}{c|}{MIT-States} & \multicolumn{4}{c|}{UT-Zappos} & \multicolumn{4}{c}{C-GQA} \\
         & METHOD & S & U & HM & AUC & S & U & HM & AUC & S & U & HM & AUC\\
        \midrule
        \multirow{10}{*}{\rotatebox[origin=c]{90}{Closed-World}} & CLIP & 30.2 & 46.0 & 26.1 & 11.0 & 15.8 & 49.1 & 15.6 & 5.0 & 7.5 & 25.0 & 8.6 & 1.4\\
        & CoOp & 34.4 & 47.6 & 29.8 & 13.5 & 52.1 & 49.3 & 34.6 & 18.8 & 20.5 & 26.8 & 17.1 & 4.4\\
        & CSP & 46.6 & 49.9 & 36.3 & 19.4 & 64.2 & 66.2 & 46.6 & 33.0 & 28.8 & 26.8 & 20.5 & 6.2\\
        & PromptCompVL & 48.5 & 47.2 & 35.3 & 18.3 & 64.4 & 64.0 & 46.1 & 32.2 & - & - & - & -\\
        & DFSP & 46.9 & 52.0 & 37.3 & 20.6 & 66.7 & 71.7 & 47.2 & 36.0 & 38.2 & 32.0 & 27.1 & 10.5\\
        & PLID & 49.7 & 52.4 & 39.2 & 22.5 & 67.3 & 68.8 & 52.4 & 38.7 & 41.0 & 38.8 & 27.9 & 11.0\\
        & Troika & 49.0 & 53.0 & 39.3 & 22.1 & 66.8 & 73.8 & 54.6 & 41.7 & 41.0 & 35.7 & 29.4 & 12.4\\
        & CDS-CZSL & 50.3 & 52.9 & 39.2 & 22.4 & 63.9 & 74.8 & 52.7 & 39.5 & 38.3 & 34.2 & 28.1 & 11.1\\
        & Retri-Aug & 50.0 & 53.3 & 39.2 & 22.5 & 69.4 & 72.8 & 56.5 & 44.5 & 45.6 & 36.0 & 32.0 & 14.4\\
        \rowcolor{gray!20}
        \cellcolor{white}& HGRL(Ours) & 51.8 & 53.0 & \textbf{40.2} & \textbf{23.3} & 73.2 & 73.6 & \textbf{59.0} & \textbf{46.8} & 44.6 & 41.9 & \textbf{34.6} & \textbf{16.3}\\
        \midrule
        \multirow{10}{*}{\rotatebox[origin=c]{90}{Open-World}} & CLIP & 30.1 & 14.3 & 12.8 & 3.0 & 15.7 & 20.6 & 11.2 & 2.2 & 7.5 & 4.6 & 4.0 & 0.3\\
        & CoOp & 34.6 & 9.3 & 12.3 & 2.8 & 52.1 & 31.5 & 28.9 & 13.2 & 21.0 & 4.6 & 5.5 & 0.7\\
        & CSP & 46.3 & 15.7 & 17.4 & 5.7 & 64.1 & 44.1 & 38.9 & 22.7 & 28.7 & 5.2 & 6.9 & 1.2\\
        & PromptCompVL & 48.5 & 16.0 & 17.7 & 6.1 & 64.6 & 44.0 & 37.1 & 21.6 & - & - & - & -\\
        & DFSP & 47.5 & 18.5 & 19.3 & 6.8 & 66.8 & 60.0 & 44.0 & 30.3 & 38.3 & 7.2 & 10.4 & 2.4\\
        & PLID & 49.1 & 18.7 & 20.4 & 7.3 & 67.6 & 55.5 & 46.6 & 30.8 & 39.1 & 7.5 & 10.6 & 2.5\\
        & Troika & 48.8 & 18.7 & 20.1 & 7.2 & 66.4 & 61.2 & 47.8 & 33.0 & 40.8 & 7.9 & 10.9 & 2.7\\
        & CDS-CZSL & 49.4 & 21.8 & 22.1 & 8.5 & 64.7 & 61.3 & 48.2 & 32.3 & 37.6 & 8.2 & 11.6 & 2.7\\
        & Retri-Aug & 49.9 & 20.1 & 21.8 & 8.2 & 69.4 & 59.4 & 47.9 & 33.3 & 45.5 & 11.2 & 14.6 & 4.4\\
        \rowcolor{gray!20}
        \cellcolor{white}& HGRL(Ours) & 51.4 & 21.1 & \textbf{22.3} & \textbf{8.7} & 73.2 & 59.6 & \textbf{51.0} & \textbf{37.5} & 44.6 & 11.8 & \textbf{15.2} & \textbf{4.5}\\
        \bottomrule
    \end{tabular}
    \centering
    \caption{Main results on there datasets. }
    \label{tab:mainresults}
\end{table*}

\noindent\textbf{Datasets and Evaluation Metrics.} We perform experiments on three commonly used datasets including MIT-States~\cite{isola2015discovering}, UT-Zappos~\cite{yu2014fine} and C-GQA~\cite{naeem2021learning}. MIT-States has 115 states and 245 objects. The number of seen compositions is 1262 and unseen compositions is 400. UT-Zappos is a small footwear dataset with 16 states and 12 objects. There are 83 seen compositions for training and 18 unseen compositions for testing. C-GQA is a challenging dataset containing 413 states and 674 objects. There are 5592 seen compositions and 923 unseen combinations. We follow the evaluation protocol of previous work\cite{huang2024troika}\cite{jing2024retrieval}. Accuracy curve is plotted by adjusting the calibration bias for the scores of unseen pairs. The area under the curve called \textbf{AUC} is recorded to measure model performance. Simultaneously the best seen accuracy \textbf{S} and the best unseen accuracy \textbf{U} are reported. We also report the best harmonic mean \textbf{HM} between seen accuracy and unseen accuracy.

\noindent\textbf{Implementation Details.} For a fair comparison, we adopt the pre-trained CLIP~\cite{radford2021learning} ViT-L/14 model as the backbone. The MLP in GAVR is one-layer Multi-head Attention and the expert is two-layer fully-connected networks. The MLP in GAPE is two-layer fully-connected networks. The learning rate is $5e-4$ for UT-Zappos and $5e-5$ for MIT-States and C-GQA. The batch size is 180 for UT-Zappos and 32 for MIT-States and C-GQA. We use Adam optimizer to train the model. The group number of state $k_s$ and object $k_o$ are set to $3$ for UT-Zappos and $5$ for MIT-States and C-GQA. The hyper-parameter $\lambda$ is set to $1.0$ for UT-Zappos and $0.1$ for MIT-States and C-GQA.

\subsection{Main Results}

For a fair comparison, we compare our methods with the state-of-the-art methods that employ the same backbone, including CLIP~\cite{radford2021learning}, CoOp\cite{zhou2022learning}, CSP~\cite{nayak2022learning}, PromptCompVL~\cite{xu2022prompting}, DFSP~\cite{lu2023decomposed}, PLID~\cite{bao2025prompting}, Troika~\cite{huang2024troika}, CDS-CZSL~\cite{li2024context} and Retri-Aug~\cite{jing2024retrieval}. The main results are presented in Table \ref{tab:mainresults}. 

We compare the results under both closed-world and open-world settings. AUC and HM are the two most important metrics. From the table, we can observe that our method beats the other methods by a significant margin. Specifically, in both AUC and HM metrics, we achieve the best results on all three datasets. In particular, we lead the second place by 2.3\% and 2.5\% in AUC and HM in the closed-world setting on UT-Zappos, and by 4.2\% and 2.8\% in AUC and HM in the open-world setting. In the closed-world setting on the challenging dataset CGQ we lead the second place by 1.9\% and 2.6\% in AUC and HM. For MIT-States, which has relatively more noise. Our method leads the second place by 0.8\% and 0.9\% in AUC and HM in the closed-world setting. These results illustrate that the proposed method effectively improves the generalization of the model.

\begin{table}[t]
    \centering
    \begin{tabular}{l | c | c c | c c}
        \toprule
         & & \multicolumn{2}{c|}{UT-Zappos} & \multicolumn{2}{c}{C-GQA}\\
         & & HM & AUC & HM & AUC\\
        \midrule
        \rowcolor{gray!20}
        \cellcolor{white}\multirow{5}{*}{\rotatebox[origin=c]{90}{CW}} &  Full & 59.0 & 46.8 & 34.6 & 16.3\\ 
        &\textit{w/o} GAVR & 52.1 & 39.4 & 28.9 & 11.7\\
        & \textit{w/o} TCPG & 55.5 & 43.1 & 32.1 & 14.4\\
        & \textit{w/o} DGP & 52.8 & 40.4 & 31.3 & 13.8\\
        & \textit{w/o} GAPE & 55.5 & 43.4 & 33.3 & 15.1\\
        \midrule
        \rowcolor{gray!20}
        \cellcolor{white}\multirow{5}{*}{\rotatebox[origin=c]{90}{OW}} &  Full & 51.0 & 37.5 & 15.2 & 4.5\\
         & \textit{w/o} GAVR & 47.0 & 33.0 & 12.5 & 3.2\\
        & \textit{w/o} TCPG & 50.1 & 36.4 & 13.8 & 3.9\\
        & \textit{w/o} DGP & 48.5 & 35.0 & 13.6 & 3.7\\
        & \textit{w/o} GAPE & 49.6 & 36.0 & 13.5 & 3.9\\
        \bottomrule
    \end{tabular}
    \centering
    \caption{Ablation studies. \textit{CW} and \textit{OW} mean closed-world and open-world. \textit{TCPG} means text co-occurrence probability graph. $w/o$ means remove.}
    \label{tab:ablation}
\end{table}

\begin{figure*}[t]
    \centering
    \begin{minipage}{\linewidth}
    \centering
    \includegraphics[width=0.9\textwidth]{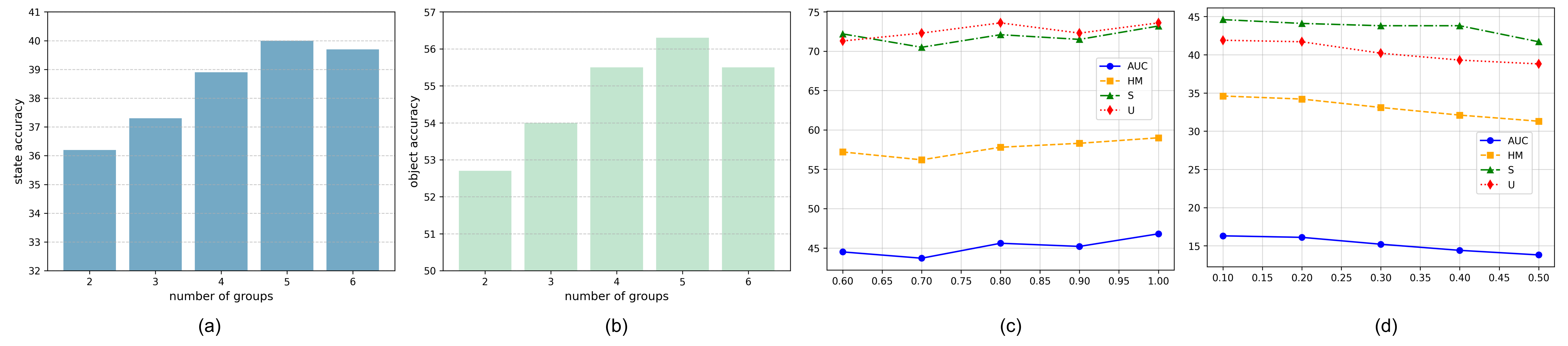}
    \end{minipage}
    \caption{(a) The effect of group number for state branch. (b) The effect of group number for object branch. (c) The sensitivity of $\lambda$ on UT-Zappos. (d) The sensitivity of $\lambda$ on C-GQA.}
    \label{fig:empir}
\end{figure*}

\begin{figure*}[htbp]
    \centering
    \begin{minipage}{\linewidth}
    \centering
    \includegraphics[width=0.9\textwidth]{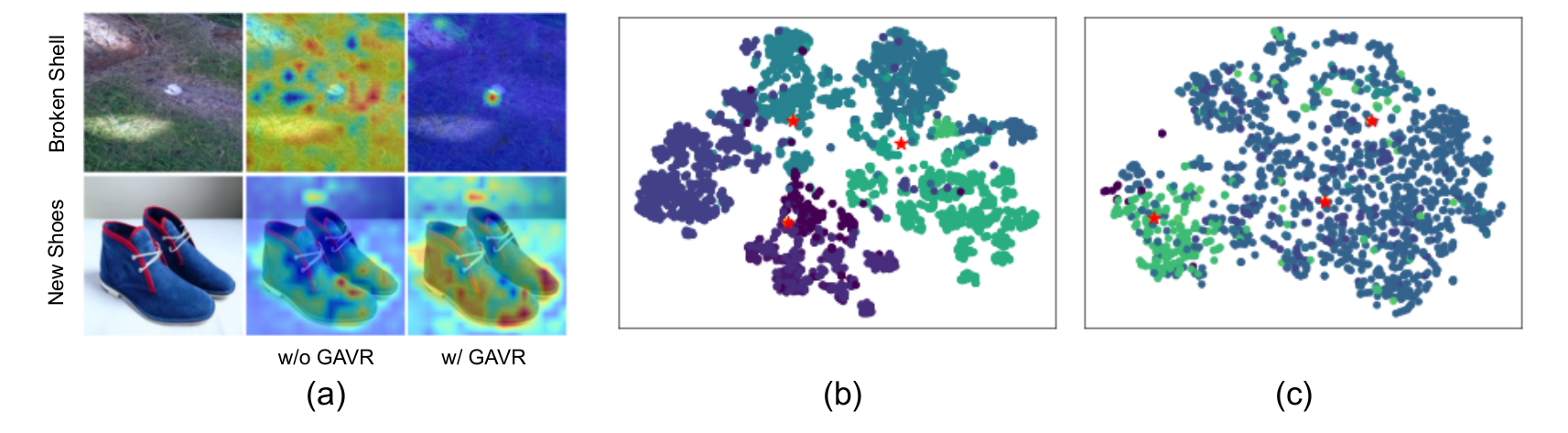}
    \end{minipage}
     \vspace{-3mm}
    \caption{(a) Attention visualization for GAVR. (b-c) T-SNE analysis for DGP. Red pentagrams indicate group prompt representations. Dots indicate image representations. (b) Shows the state whose label is \textit{Synthetic} in UT-Zappos. Different colored dots indicate different objects. (c) Shows the object whose label is \textit{Shoes.Sneakers.and.Athletic.Shoes} in UT-Zappos. Different colored dots indicate different states.}
    \label{fig:attention}
\end{figure*}

\subsection{Ablation Study}

We conduct extensive empirical analysis to assess the impact of the individual modules as well as the hyper-parameters.

\textbf{The effect of key modules.} A series of ablation experiments are conducted to investigate the role of each module. The results of the experiments are shown in Table \ref{tab:ablation}. From the table, it can be observed that removing any of the modules degrades the performance of the model, indicating the rationality of the design of the proposed method. Among them, GAVR and DGP have the greatest impact on the model's performance, indicating that they contribute greatly to the optimization of the visual and prompt representations, respectively. In addition, the results show that the introduction of the text co-occurrence probability graph (TCPG) improves the performance of the model, demonstrating that it indirectly facilitates the representational ability of GAVR, guiding it to discover and aggregate homogeneous groups.

\textbf{The effect of group number $k_s$ and $k_o$.} Due to the lack of supervision by hierarchical labels, the potential number of groups is agnostic. We investigate the relationship between the setting of the group number and the model performance. The results are shown in Fig. \ref{fig:empir} (a) and (b), which demonstrates the effect of the setting of the group number on the state accuracy and object accuracy on MIT-States. The results show that when the number of groups is set too small, e.g., 2, a substantial performance decay occurs, indicating that the weak expert panel has difficulty in handling the heterogeneous group problem. And when the number of groups gradually increases, the performance does not increase linearly, which may be caused by the saturation of the number of groups.

\textbf{The effect of hyper-parameter $\lambda$.}  We investigate the effect of the coefficient $\lambda$ in Eq. \ref{Eq_total_loss} on the performance of the model. The results are shown in Fig. \ref{fig:empir} (c) and (d). It can be observed that the sensitivity of the model to $\lambda$ is within a reasonable range. We suggest setting a larger value on a small dataset such as UT-Zappos to promote fast model fitting, and a smaller value on a large dataset such as C-GQA to avoid unstable training.

\subsection{Visualization Analysis}

To deeply investigate the role of the key components, we perform a series of visualization analyses.

\textbf{Attention visualization for GAVR.} To further explore the effect of GAVR, we perform a visual comparison. As can be seen in Fig. \ref{fig:attention} (a), single network ($w/o$ GAVR) is driven to learn semantic information shared between heterogeneous groups thus easily capture pseudo-related features. In contrast, GAVR has multiple expert sub-networks that can adaptively process data from different groups, and thus the captured features are more accurate (top row). Meanwhile, the expert networks complements the group-specific semantic knowledge, so the extracted features are richer and more discriminative (bottom row).

\textbf{T-SNE Analysis for DGP.} To study the context learned by group prompts, we map them to the visual space. Fig. \ref{fig:attention} (b) and (c) show the feature distribution of the same state across different objects and the same object across different states, respectively. It can be observed that the significant group heterogeneity in the same state (object). Meanwhile, The proposed group prompts capture different contexts. These group prompts provide distributed centers so that visual features do not have to be compressed into a single prototype thereby maintaining group-specific semantic information.

\section{Conclusion}

In this paper, we explore the drawback of existing approaches to deal with conditional dependencies, i.e., the zero-sum game of transferability and discriminability. To solve this problem, we provide a new perspective, inspired by the human notion of hierarchical clustering, that formulates state (object) representation learning into homogeneous group representation learning. To this end, we propose a new framework, HGRL, which aims to adaptively discover homogeneous groups and extract group-specific semantic information. Our method consists of three key components to facilitate both visual and prompt representations. Experiments demonstrate the effectiveness of the proposed method.

\bibliographystyle{named}
\bibliography{ijcai25}

\end{document}